\pdfoutput=1

\documentclass[11pt]{article}

\usepackage[]{ACL2023}

\usepackage{times}
\usepackage{latexsym}
\usepackage{graphicx}
\usepackage{comment}

\usepackage[T1]{fontenc}

\usepackage[utf8]{inputenc}

\usepackage{microtype}

\usepackage{inconsolata}

\title{Don't Believe Everything You Read: Enhancing Summarization Interpretability through Automatic Identification of Hallucinations in Large Language Models}

\author{Priyesh Vakharia\textsuperscript{†}\thanks{\quad  These authors contributed equally to this work.} \and  
Devavrat Joshi\textsuperscript{†}\footnotemark[1] \and Meenal Chavan\textsuperscript{†}\footnotemark[1] \AND
Dhananjay Sonawane\textsuperscript{†}\footnotemark[1] \and
Bhrigu Garg\textsuperscript{†}\footnotemark[1] \and
Parsa Mazaheri\textsuperscript{†}\footnotemark[1] \\
\textsuperscript{†}University of California Santa Cruz \\
\texttt{\{pvakhari, dsjoshi, mmchavan, dsonawan, bgarg, pmazaher\}@ucsc.edu}
}

\begin{document}

\maketitle
\begin{abstract}
Large Language Models (LLMs) are adept at text manipulation -- tasks such as machine translation and text summarization. However, these models can also be prone to hallucination, which can be detrimental to the faithfulness of any answers that the model provides. Recent works in combating hallucinations in LLMs deal with identifying hallucinated sentences and categorizing the different ways in which models hallucinate. This paper takes a deep dive into LLM behavior with respect to hallucinations, defines a token-level approach to identifying different kinds of hallucinations, and further utilizes this token-level tagging to improve the interpretability and faithfulness of LLMs in dialogue summarization tasks. Through this, the paper presents a new, enhanced dataset and a new training paradigm.
\end{abstract}

\section{Introduction}

The primary capability of most state-of-the-art LLMs is their ability to generate text, encapsulating tasks such as machine translation and text summarization. Text summarization, in particular, can be complex; the task requires fetching all possible information from prose, understanding the most important bits, and placing them all together in a precise manner. Depending on the type of text that the model is presented with, (long scientific texts, articles, dialogues), the exact goal of the task also changes to some extent. While strides in LLM research have seen great improvements in a model's ability to successfully summarize text, they are still prone to hallucination. Hallucination not only reduces the accuracy of a model, it also makes the model unfaithful and unreliable. Thus, hallucination identification and correction are of great importance in improving LLM performance and reliability.
\begin{figure}[h]
  \includegraphics[width=0.5\textwidth]{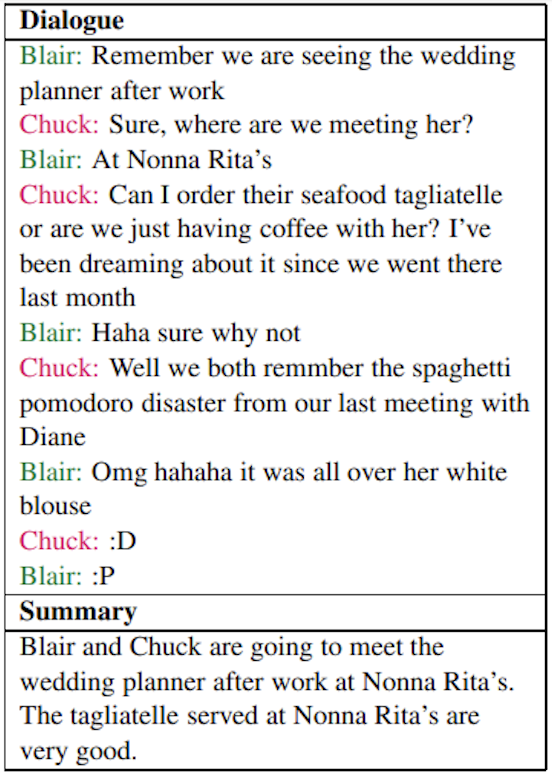}
  \caption{ Example of dialogue from the SAMSum corpus.}
  \label{dg}
\end{figure}
To correct hallucinations, there first needs to be a proven method of identifying and assessing these hallucinated outputs. Previous works have shown that the currently existing metrics for text evaluation such as BLEU Scores\cite{papineni2002bleu}, ROUGE\cite{lin-2004-rouge}, etc, do not correlate with the faithfulness of the model outputs\cite{maynez2020faithfulness}.
More recently, efforts have begun trying to develop automatic ways to identify and assess hallucinated output\cite{ji2023survey}. \cite{zhou2021detecting} builds a classifier model that predicts whether a token in the summary is hallucinated or not.
However, if these approaches are to be proposed as metrics, then they need to be interpretable and
faithful. According to \cite{lyu2023faithful}, faithfulness refers to the extent to which an explanation accurately reflects.
Interpretability, on the other hand, stems from a balance between faithfulness and plausibility, where plausibility
refers to how convincing is the explanation to humans \cite{jacovi2020faithfully}. In other words, an interpretation of a
model must be faithful to the true reasoning process as well as plausible to humans.

This project takes a deep dive into two different approaches for identifying hallucination in model-generated summaries and using this understanding of model behavior to improve the interpretability and faithfulness of generated summaries. 

Current research in hallucination detection involves either binary classification of hallucination (whether a summary is hallucinated or not) [1] or breaks down hallucinations in summaries into different types on a summary level. For this project, the goal was to detect hallucination on a token level. To achieve this, human annotation was performed, specifically on the ConFiT dataset (based on the SAMSUM dataset) to create a more enhanced token-level annotated SAMSUM dataset.

We use this unique dataset in three different ways:
\begin{enumerate}
    \item In section \ref{approach1}, we train a proxy "explainer" model to take in a set of dialogues and a corresponding summary and label each token with a "faithfulness" tag.
    \item In section \ref{joint}, we build upon the lack of the faithfulness of using a proxy model \cite{lyu2023faithful} by combining the tasks of summarizing and faithfulness tagging.
    \item In section \ref{llm-experiments}, we experiment with different kinds of prompting techniques to produce better summaries using state-of-the-art LLMs. These techniques attempt to take into account the chance of hallucination on a token-level and chain-of-thought prompting to improve summarization performance.   
\end{enumerate}

\section{Data}

The dataset utilized in this project is the SAMSum dataset \cite{gliwa-etal-2019-samsum}, which comprises approximately 16,000 messenger-like conversations along with corresponding summaries. These conversations were meticulously crafted and transcribed by linguists fluent in English. The linguists were instructed to create conversations that resemble their daily written interactions, capturing the topic proportions found in their real-life messenger conversations. Consequently, the dataset exhibits a diverse range of styles and registers, encompassing informal, semi-formal, and formal conversations that may incorporate elements such as slang words, emoticons, and typographical errors. To facilitate analysis, the conversations were further annotated with summaries aiming to provide a concise overview of the conversation's content in the third-person perspective. A snapshot of the dataset can be seen in Figure \ref{dg}.

\begin{figure*}[h]
  \centering
  \includegraphics[width=14cm, height=6cm]{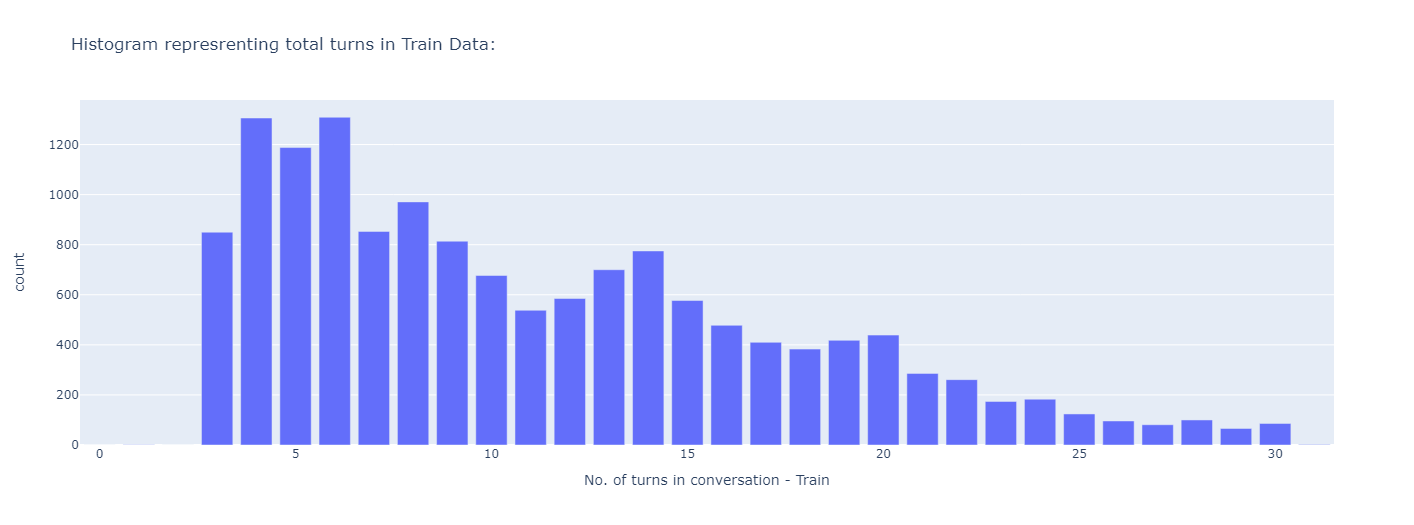}
  \caption{Distribution of total turns in the train data}
  \label{turns}
\end{figure*}

The dataset is evenly divided into four categories depending on the number of utterances within each conversation: 3-6, 7-12, 13-18, and 19-30.  Figure 2 shows this distribution over the train data. Each utterance within the dataset includes the name of the speaker. Approximately 75 percent of the conversations predominantly involve dialogues between two participants, while the remaining conversations involve interactions among three or more individuals. On performing some additional exploratory data analysis, it was also found that the length of summaries of the train data ranged from ten to twenty words as shown in Figure \ref{turns}.

While the SAMSum dataset has gold summaries, it lacks any form of hallucinations within those summaries, since they were manually generated by a team of linguists. For this project, we place emphasis on the modified SAMSum dataset used in CONFIT \cite{tang-etal-2022-confit}. CONFIT is a training strategy specifically designed to enhance the factual consistency and overall quality of summaries using a novel contrastive fine-tuning approach. Additionally, in order to gain a deeper understanding of the hallucinations generated by pre-trained models, the researchers in CONFIT developed a linguistically motivated taxonomy of hallucination errors for dialogue summarization. This taxonomy goes beyond a binary classification of faithful or unfaithful summaries and provides a more nuanced categorization of the types of factual errors present in summaries. 

\begin{figure*}[h]
  \centering
  \includegraphics[width=14cm, height=6cm]{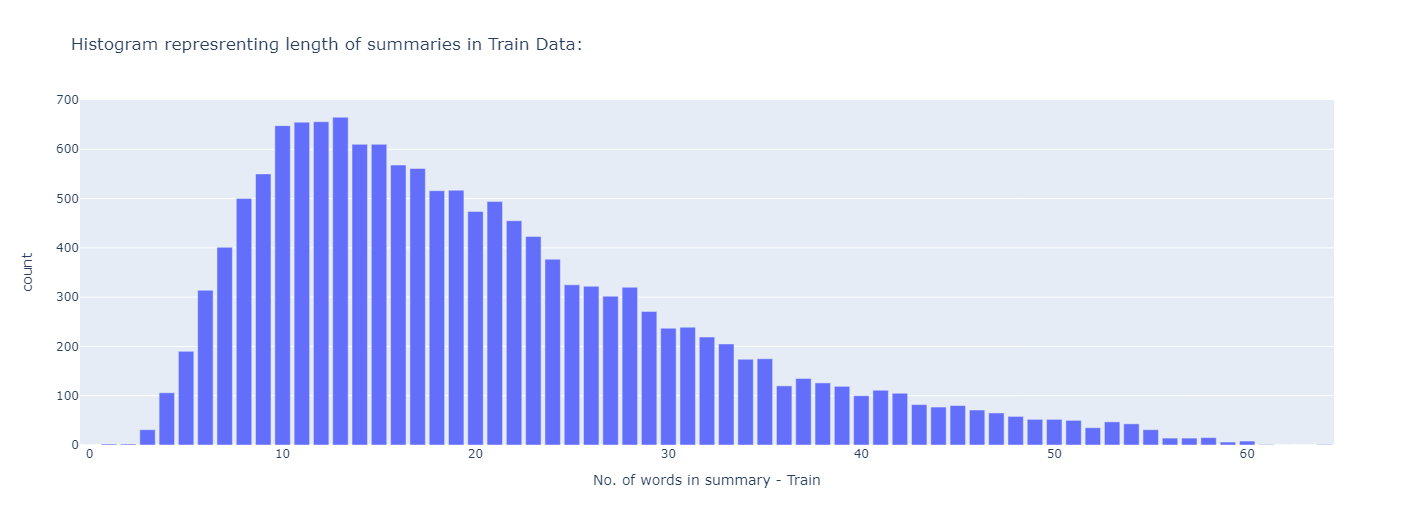}
  \caption{Length of summaries in the train data}
  \label{haldist}
\end{figure*}

\subsection{Data annotation and guidelines}
\label{hallucinationcat}

In the annotated version of SAMSum used in CONFIT, each generated summary is labeled with the specific type of error it contains. The objective is to leverage these identified error types and apply the token-level annotation to the data following the provided guidelines. This annotation process allows for errors to be marked at a more granular level, providing deeper insights into the distribution and characteristics of errors within the dataset. 

The faithfulness tags are defined according to the hallucination categories below -- 

1. Wrong Reference Error- A pronoun in the generated summary either refers to the wrong or non-existent noun it should be replaced, or when a personal named entity in the summary is used incorrectly instead of a different personal entity mentioned in the reference.

Example: [Reference Summary] Mohit asked Darlene about the test.
[Model-Generated Summary] Darlene asked Mohit about the test.

2. Object Error- Factual errors that arise from inaccuracies in either the direct or indirect objects. 

Example: [Reference Summary] Tara raised her glass.
[Model-Generated Summary] Tara raised her spoon.

3. Circumstantial Error: Circumstantial information (e.g., date, time, location) about the predicate doesn’t match the reference.

Example: [Reference Summary] The USA was founded in 1776.
[Model-Generated Summary] The USA was founded in 1767.

4. Other Uncommon Errors - Errors that encompass factual errors resulting from discrepancies in grammatical tense between the generated summary and the reference.

Example: [Reference Summary] The children will go to the library.
[Model-Generated Summary] The children went to the library.

5. Not Hallucinated- Tokens in the summary that are not hallucinated.

6. Missing Information- A special tag to be given at the end of sentence token to indicate if a summary suffers from missing information hallucination.

\subsection{Data Analysis}

\begin{table}[htp]
    \centering
    \begin{tabular}{ |c|c| } 
    \hline
        Error Tag & Description \\
        \hline\hline
        W & Wrong reference error \\
        OB & Object error \\
        C & Circumstantial error \\
        N & Other uncommon errors \\
        O & Not hallucinated \\
        M & Missing information \\
    \hline
    \end{tabular}
    \caption{Hallucination categories}
    \label{table:categories}
\end{table}

Table \ref{table:categories} shows the faithfulness tags we assign to each category listed above. Every token is labeled with the faithfulness tag corresponding to the category they fall in as defined above.

\begin{figure}[h]
  \centering
  \includegraphics[width=0.5\textwidth]{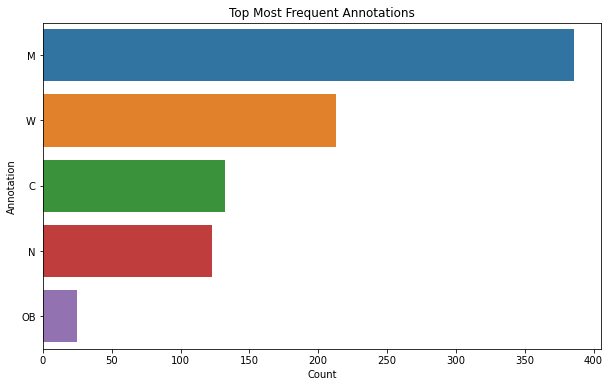}
  \caption{Frequency of Most Common Errors}
  \label{fig:Frequency of Most Common Errors}
\end{figure}

Six human annotators meticulously annotated 600 conversations with token-level granularity. Upon completion, the resulting analysis revealed that 92.6\% of the 10930 tokens remained unaltered, as depicted in Figure 4. Figure 4 visually represents the distribution of errors at the token level. Predominantly, missing information emerged as the most frequent error, occurring 386 times, constituting 3.3\% of the tokens. Following closely behind, the wrong reference error occurred 212 times. Additionally, circumstantial errors surfaced in 131 instances, while other uncommon errors occurred 124 times. Notably, object errors proved to be the rarest, presenting a total of 25 occurrences.

Since \cite{zhou2020detecting} has many hallucinated summaries for each dialog, the test and validation set are prepared by identifying unique dialogues and placing all summaries associated with a dialogue in a single split. Accordingly, for working with this dataset, the distribution for train, validation, and test is (76, 12, 12). \cite{zhou2020detecting} employed five distinct models for summarization: BART, BART-CONFIT, Pegasus, Pegasus-CONFIT, T5, and T5-CONFIT. Figure 5 shows the distribution of all the labels for each model. During an analysis of the summary tokens produced by these models, a pattern emerged regarding the sources of various errors. Specifically, Pegasus exhibited a higher occurrence of circumstantial errors. T5, on the other hand, demonstrated a tendency to omit information more frequently. Notably, the occurrence of uncommon errors was most prevalent within the outputs of BART. Interestingly, Pegasus presented the least number of object errors, while BART stood out as primarily responsible for wrong reference errors as represented in Figure 5.

\begin{figure*}[h]
  \centering
  \includegraphics[width=12cm]{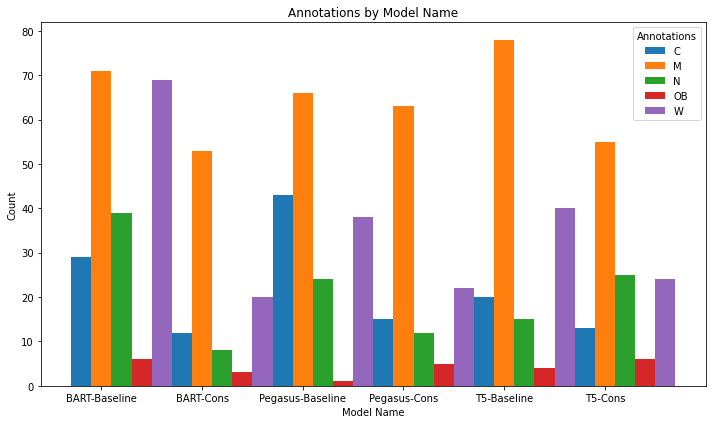}
  \caption{Frequency of Most Common Errors per Model}
  \label{fig:Frequency of Most Common Errorss per Model}
\end{figure*}

\section{Proxy Model for Hallucination Tagging}
\label{approach1}

For the first approach, the aim is to classify the hallucinated tokens based on Table 1. Previous work only classified tokens in a binary fashion based on whether they were hallucinated or not. The proxy mode builds on this by identifying various types of hallucination at a token level. Such information aids in interpretability and can help in deciding which types of hallucinations to prioritize in terms of reduction when creating a model.

\subsection{Baseline}

For this approach, \cite{zhou2020detecting} for token-level hallucination detection will serve as a baseline model. This baseline uses binary hallucination detection, as opposed to the proposed approach, which aims for multiclass token-level hallucination detection. In order to employ this baseline in a multiclass context, the gold faithfulness tags are converted to a binary set, mapping all non-O tags to 1s and O tags to 0s. This baseline can be evaluated both with and without gold summaries provided as context for tagging reference summaries.

\subsection{Novelty}

Previous work focuses on performing a binary classification on tokens in the summary and predicting if a token is hallucinated or not. This is difficult to interpret since it provides no reasoning for why the model deemed a token hallucinated. The annotated tags in this enhanced dataset are generally correlated with certain functions in a sentence and it is more sound to infer what type of hallucination the model believes the token to be.

The proposed approach has an added interpretability factor by assigning each token with a certain hallucination category or as defined above a "faithfulness" tag. This helps in a better interpretation of the reasoning of the model when it classifies a certain token as hallucinated or not.

\subsubsection{Training Setup}

For this approach, a seq2seq transformer model\cite{vaswani2017attention} is used alongside GPT4 to derive hallucination tags. The encoder-only transformer is going to take the dialogue and the summary as the input and predict the faithfulness tag for each token in the summary. 

Further, the BigBird model \cite{zaheer2020big} is leveraged for the task of token classification in a summary because it is the only publicly encoder-only model on HuggingFace that can handle more than 512 tokens, which account for 11\% of the dataset's summaries. By fine-tuning BigBird on this token classification task, the aim is to leverage its contextualized representations and transformer-based architecture to accurately detect and classify token-level hallucinations in a summary. 

Two types of training are presented, hallucinated summary only training (HS) and training with gold summary (GS).

\begin{figure*}[h]
  \centering
  \includegraphics[width=11cm]{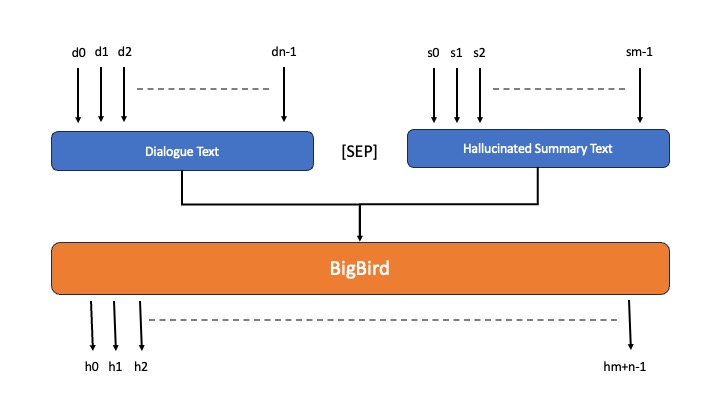}
  \caption{Model Design for Approach 1 without Gold Summaries}
  \label{bbwg}
\end{figure*}

For HS, the dialogue and the hallucinated summary (\emph{L, S}) are concatenated, separated with a separator token, [SEP] and with an [EOS] at the end of the summary in order to tag the missing information, as a single input sequence to the model.

\begin{figure*}[h]
  \centering
  \includegraphics[width=11cm]{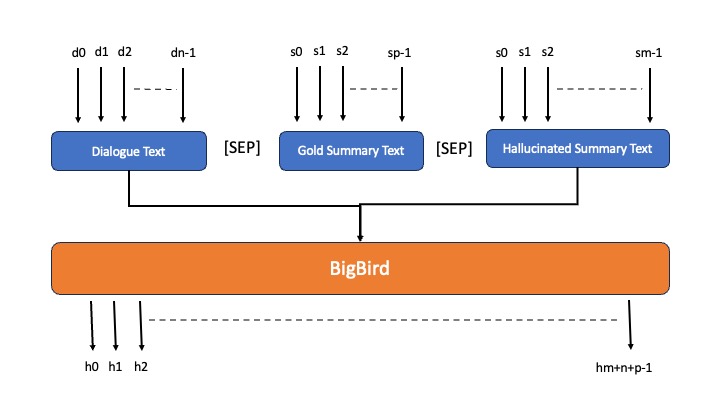}
  \caption{Model Design for Approach 1 without Gold Summaries}
  \label{bbg}
\end{figure*}

Then, the standard classification loss, $L_{pred}$ is minimized over the psuedo "faithfulness" tags $L_s$ on top of the final hidden vectors of each token in \emph{S}, as shown in Figure \ref{bbwg} for hallucinated summaries only training and in Figure \ref{bbg} for training with the gold summaries.

For GS, the dialogue, the gold summary, and the hallucinated summary (\emph{L, G, S}) are concatenated, separated with a separator token, [SEP] between each, and with an [EOS] at the end of the hallucinated summary.

\subsection{Experiment Plan}

\subsubsection{Fine-tuning}

The experiment plan involves fine-tuning the language model using token-level annotated CONFIT data. Table \ref{table:hyperpram} list downs the hyperparameters for the HS and the GS. 

\begin{table}[htp]
    \centering
    \begin{tabular}{ |c|c| } 
    \hline
    learning rate & 2e-5\\
    batch size & 1\\
    num train epochs & 10\\
    weight decay & 0.01\\
    metric for best model & "f1"\\   
    \hline
    \end{tabular}
    \caption{Hyperparameters}
    \label{table:hyperpram}
\end{table}

The model will tag the entire input but only the summary tags will be considered in evaluation.

The losses for finetuning with GS and HS are similar but GS is slightly lower.

\subsubsection{GPT4 Few Shot Tagging}

The second proxy approach involves using GPT4 with various prompts to tag a summary given its dialogue and summary. 

\subsection{Evaluation}

Given the gold set of "faithfulness" tags for each summary, and the predicted tags, the accuracy, precision, recall, and F1 score for the set of tags are evaluated in the style of \emph{seqeval}.

\subsection{Results}

\begin{table*}[htp]
    \centering
    \begin{tabular}{ |c|c|c|c|c| } 
    \hline
        & Precision & Recall & F1 & Accuracy \\
        \hline\hline
        Baseline(HS) &  0.06991 & 0.40566 & 0.11927 & 0.73240\\
        Baseline(GS) &  0.08641 & 0.39622 & 0.14189 & 0.75916\\
        Fine Tuning(HS) & 0.66667 & 0.72727 &
        0.69565 & 0.99597\\
        Fine Tuning(GS) & 0.66667 & 0.72727 &
        0.69565 & 0.99651\\
        GPT4 (HS) Prompt 2 & 0.27272 & 0.54545 & 0.36363 & 0.8928\\
        GPT4 (HS) Prompt 3 &  0.26829 & 1.0 & 0.42307 & 0.84693\\
        GPT4 (HS) Prompt 6 & 0.42857 & 0.81818 &  0.5625 & 0.92857\\
        GPT4 (HS) Prompt 8 & 0.53846 & 0.63636 & 0.58333 & 0.94897 \\
        GPT4 (HS) Prompt 9 & 0.58823 & 0.90909 & 0.71428 & 0.95918\\
    \hline
    \end{tabular}
    \caption{Proxy Tagging Results Table}
    \label{table:tag_res}
\end{table*}

Table \ref{table:tag_res} denotes the precision, recall, f1, and accuracy scores for all the experimentation. The first prompt, placed in the system prompt, simply contained: the names of the available tags, one example of a summary, dialogue, and lastly a string of tags. It was observed that the predicted tag length seldom matched the gold tags, as the model struggled to keep count of the word number even when properly formatted output was produced. Only 10\% of the data produced valid output for which F1 could be calculated.

In Prompt 2, when describing tagging, the tags were placed next to each associated word and M was placed next to [EOS]. This fixed the length discrepancies between gold and predicted tags but the model still made mistakes when placing the M tag. Two examples were given. This represented the first real F1 result of 0.36.

Prompt 3 gave 3 more informative examples covering more tags, made tagging a 2 step process of first tagging existing tokens and then checking for missing information, with human-provided explanations for both steps. Additionally, each complete example was moved into the chat history. This combination improved performance by 6 points.

In Prompt 6, the M tag was removed from the system prompt description. The explanations were also altered to be more conducive to token-level tagging and descriptions of each tag were added. These descriptions led to a further 12 point F1 increase.

After attempting to refine the tagging explanations in prompts 3-7, the Prompt 6 explanations were removed in Prompt 8, improving performance by 2 points. Providing explanations with descriptions was found to hurt model performance.

In prompt 9, the tag descriptions were edited to make them more precise. In summary, prompt 9 only contained task and tag descriptions in the system prompt and output examples with tags printed next to tokens in the chat history. The final F1 was 0.71.

All prompts can be seen in the appendix with the best prompt, prompt 9, appended to the top of the tagging section within the appendix.

\begin{figure*}[h]
  \centering
  \includegraphics[width=9cm]{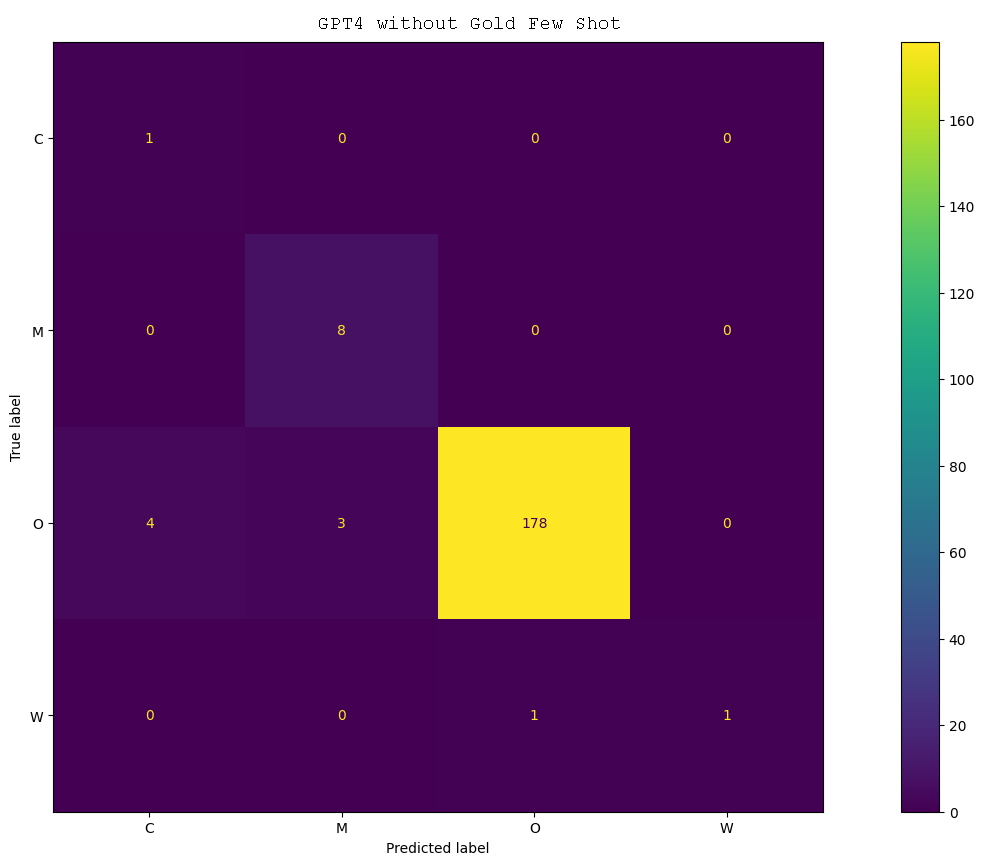}
  \caption{Confusion matrix for GPT4 without gold summary few-shot inferences}
  \label{GPT4CM}
\end{figure*}

The baseline performs poorly but the model after fine-tuning performs quite well. However, adding the gold summaries during training does not provide any statistically significant improvement to training without the gold summary. With training, it is seen that the model does not capture trends for any tag except O and M as shown in Figures \ref{WGCM} and \ref{GCM} in the Appendix. GPT tagging was better than training but much slower at scale and more time-intensive for prompt tuning. In the GPT experiments, it was found that asking it to generate tags alongside given summary tokens and then separately consider the Missing Information tag was the best approach. Additionally, explanations often misled the model and more detailed descriptions were more helpful than explanations. In the best prompt, the most frequent error was confusing the W and C tags, according to Figure \ref{GPT4CM}. 

\section{Experiments with Large Language Models}
\label{llm-experiments}

\subsection{Goal}

The objective of this approach is to use different ways of prompting in state-of-the-art LLMs (GPT-4, LLaMa 2). This is to understand whether a model performs better if it is prompted to take hallucinations into account in the newly generated summary.

\subsection{Prompt Tuning}

\subsubsection{Baseline Prompt}

The baseline prompt serves as the backbone for this task. The task described is a relatively straightforward dialogue summarization task. \\

\textbf{Sample Prompt for Baseline Prompting}
\begin{quote}
    Generate a summary of a length of exactly 10 to 15 words for the given set of dialogues.

    Dialogue:
\begin{quote}
        "Hannah: Hey, do you have Betty's number? \\
    Amanda: Lemme check \\
    Hannah: GIF \\
    Amanda: Sorry, can't find it. \\
    Amanda: Ask Larry \\
    Amanda: He called her last time we were at the park together \\
    Hannah: I don't know him well \\
    Hannah: GIF \\
    Amanda: Don't be shy, he's very nice \\
    Hannah: If you say so.. \\
    Hannah: I'd rather you texted him \\
    Amanda: Just text him \\
    Hannah: Urgh.. Alright \\
    Hannah: Bye \\
    Amanda: Bye bye"
\end{quote}    
    Summary: 
\end{quote}

This backbone prompt was then modified into zero-shot, one-shot, and few-shot variants using fewer or more examples picked from the SAMSum dataset.

\subsubsection{Hallucination-based Prompting}

To improve upon the baseline prompt, the model was asked to do token-level tagging on the generated summary based on whether it was hallucinated or not. \\

\textbf{Sample Prompt for Hallucination-based Prompting}
\begin{quote}
    Given a set of dialogues, the task is to generate a summary of 10-15 words by considering all the dialogues, and perform token-level classification on the summary based on whether it is hallucinated or not. Use the following tag classes to label each token of the summary. 
    O = Not Hallucinated,
    W =  Wrong person reference,
    C = Circumstantial error,
    OB = Object error,
    N = uncommon errors like tense errors 
    M = Missing information.
    The tag M should only be added at the end of the sequence incase the summary is missing any information and not as a tag specific to a word in the summary. 
    
    Dialogue- 
    \begin{quote}
    "Hannah: Hey, do you have Betty's number? \\
    Amanda: Lemme check \\
    Hannah: GIF \\
    Amanda: Sorry, can't find it. \\
    Amanda: Ask Larry \\
    Amanda: He called her last time we were at the park together \\
    Hannah: I don't know him well \\
    Hannah: GIF \\
    Amanda: Don't be shy, he's very nice \\
    Hannah: If you say so.. \\
    Hannah: I'd rather you texted him \\
    Amanda: Just text him \\
    Hannah: Urgh.. Alright \\
    Hannah: Bye \\
    Amanda: Bye bye"        
    \end{quote}
    
    Summary- "Amanda can't find Betty's number. Larry called her last time they were at the park together. Amanda will text Larry."
    
    Tags- "O O O O O O O O O O O O O O O O O O W O O O O"
    
    Explanation - Let's think step by step. The dialogue is about Hannah asking for Betty's number to Amanda, who couldn't find it and suggests to ask Larry for it since he had called her(Betty) the last time they were in the park together. Hannah doesn't know him(Larry) well and is shy to text him, but Amanda asks her to do it anyway.  So according to the summary, "Amanda will text Larry" is incorrect. The way to correct this information is the token Amanda can be changed to Hannah. This is Wrong Reference (W) from the tokens described above. All other tokens are correct and are thus Not Hallucinated (O).
    
    Similarly, for the next dialogue, generate summary of all the dialogues and tags for the summary. Think step by step to explain it.
    
    Dialogue- 
    \begin{quote}
    "Harry: and? have you listened to it? \\
    Jacob: listened to what? \\
    Harry: to the song i sent you 3 days ago -.- \\
    Jacob: oh shit, i completely forgot... \\
    Harry: ofc again \\
    Jacob: don't be like this :* i'll do that later tonight \\
    Harry: heh, okay \\
    Harry: i'm really curious what you'll think about it \\
    Jacob: i'll let you know, a bit busy right now, speak to you later! \\
    Harry: okay"        
    \end{quote}
    
    Summary- \\
    Tags- \\
    Explanation- \\
\end{quote}
This prompt was used on GPT-4 and LLaMa 2 in two formats: one containing the "explanation" section, which uses a chain-of-thought protocol to go step by step and explain how the model came to determine hallucinations in the generated summary, and one that doesn't contain this explanation section.

\subsection{Prompting on Fine-tuned LLMs}

To improve the performance of LLMs on the summarization task, LLaMa 2 was fine-tuned on the token-level annotated version of the SAMSum dataset described in the Data section. The fine-tuning was done using Quantized Low-Rank Adaptation of Large Language Models (QLoRA), a lightweight training technique that significantly reduces the number of trainable parameters by introducing several smaller weights in a transformer model, and those are the only weights that are trained.

The exploration encompassed LLaMa2 versions 7B, 13B, and 70B and all versions were sourced from Huggingface. Notably, due to constraints related to GPU capabilities, this model underwent quantization to 4 bits, ensuring feasibility within the computational framework.

The model was fine-tuned using a single Nvidia A100 GPU with 80GB VRAM for 5000 steps on a set of 400 samples from the annotated SAMSum dataset described in the data section. The recommended hyperparameters for LoRA fine-tuning were utilized for this. The loss curve for the fine-tuning is presented in Figure~\ref{fig:Train loss per Step}. The loss curves for 7B and 13B version can be found in Appendix~\ref{llm train loss}.

\begin{figure*}[htp]
  \centering
  \includegraphics[width=12cm]{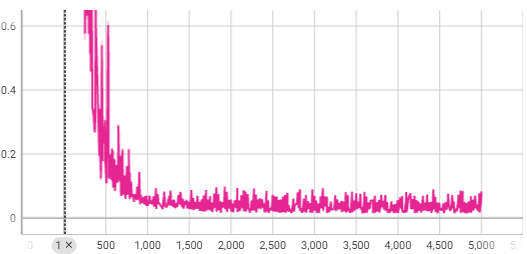}
  \caption{Train Loss per Step}
  \label{fig:Train loss per Step}
\end{figure*}

\subsection{Results and Discussion}

The performance of GPT-4 and LLaMa 2 (both the base model and the fine-tuned variant) in terms of ROUGE-1, ROUGE-2, ROUGE-L and ROUGE-Lsum scores is presented in Table~\ref{prompting-results}. The ROUGE scores are calculated using the gold summaries from the SAMSum dataset. Based on these results, it was found that models generated better summaries when they were asked to generate faithfulness tags for the summaries that they generated as per the defined rules. That said, adding a chain-of-thought protocol by asking it to generate an explanation for the tagging process didn't cause a significant improvement in performance. However, as chain-of-thought prompting usually goes, explanations tend to make answers more interpretable and, hence more reliable. 

While there was no significant change to the model performance after fine-tuning, a lot of that can be attributed to 2 factors:
\begin{quote}
1. Lack of a big dataset for fine-tuning. \\
2. Possible loss of precision due to 4-bit quantization.    
\end{quote}
Alongside that, the smaller variants of the LLaMa 2 model (7b and 13b) would not generate summaries at all as shown in Appendix~\ref{LLaMaoutput}; only the 70b model seemed to generate summaries properly.

This study shows that the newly devised method of incorporating hallucination detection into the thought process of the LLM improves its performance in a given task.

\begin{table*}[]
\centering
\begin{tabular}{|l|l|l|l|l|}
\hline
                   & ROUGE-1 & ROUGE-2 & ROUGE-L & ROUGE-Lsum \\ \hline
LLaMa2                                   & 0.4     & 0.15    & 0.31    & 0.31       \\ 
LLaMa2 Finetuned                         & 0.4     & 0.16    & 0.31    & 0.32       \\ 
Zero-Shot Summary                        & 0.39    & 0.15    & 0.32    & 0.32       \\ 
One-Shot Summary                         & 0.4     & 0.13    & 0.32    & 0.32       \\ 
One-Shot Summary + Tagging               & 0.469   & 0.184   & 0.359   & 0.36       \\
One-Shot Summary + Tagging + Explanation & 0.46    & 0.19    & 0.35    & 0.36       \\ \hline
\end{tabular}
\caption{Performance of state-of-the-art LLMs across various prompt tuning methods for summarization task.}
\label{prompting-results}
\end{table*}

\section{Joint Summarization and Tagging}
\label{joint}

\subsection{Goal}

Previous works use the proxy model to try and imitate/explain the reasoning followed by the original model. However, \cite{lyu2023faithful} casts doubt on the ability of the proxy model to accurately reflect the reasoning. 
Hence, this paper proposes merging the two tasks of summarizing and hallucination tagging into one joint task using only one model. 

\begin{figure}[h]
  \centering
  \includegraphics[width=6cm]{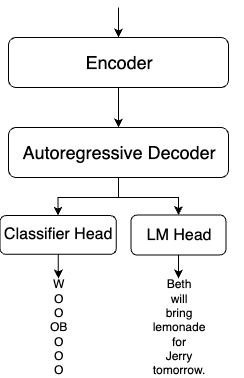}
  \caption{Joint Model}
  \label{joint_model}
\end{figure}

Figure \ref{joint_model} illustrates the novel approach presented, where the tasks of summarizing and hallucination detection are integrated into a single model. This integration is pivotal for achieving more consistent and reliable outputs. The rationale behind this design is grounded in two main hypotheses --
\begin{enumerate}
    \item \textbf{Consistency in Reasoning}: By positioning the classification layer adjacent to the decoding head, the model can simultaneously generate tags and tokens. This concurrent processing ensures that the reasoning behind both hallucination detection and summary generation is aligned, leading to outputs that are both coherent and faithful to the original content.
    \item \textbf{Enhanced Hallucination Detection}: The unified architecture allows for the direct application of hallucination detection during the summary generation process. This integration is expected to significantly improve the model's ability to identify and avoid hallucinations, thereby producing summaries that are more accurate and reflective of the original content.
\end{enumerate}

To empirically validate the proposed approach, the following hypotheses can be tested --
\begin{enumerate}
    \item Integrating summary generation with hallucination detection in a single model will lead to superior hallucination detection. This integration should result in summaries that are not only more accurate but also more faithful to the original content.
    
    \item The adoption of a joint loss function, tailored to address both summarization and hallucination detection, will align the model's training objectives. This alignment is hypothesized to minimize the occurrence of hallucinations, thereby yielding summaries that are more true to the source material.
\end{enumerate}

\subsection{Baseline}
\label{apprch2-baselines}

The BART model, as outlined in \cite{lewis2019bart}, is utilized as the foundational model for the approach presented. This choice is due to BART's proven efficacy in producing coherent and contextually relevant summaries, especially when fine-tuned on specific datasets such as SAMSum.

BART's architecture combines bidirectional and auto-regressive transformers, providing a robust framework for understanding and generating natural language. By fine-tuning BART on the SAMSum dataset, the advanced capabilities of BART in capturing dialog nuances are leveraged, making it an ideal starting point for the proposed enhancements.

\subsection{Novelty}

Previous works that involve predicting and emitting a human readable explanation perform these tasks sequentially. \cite{camburu2018esnli} uses a predict-then-explain technique, where the model first predicts then a proxy model generates an explanation for the prediction. On the other hand, \cite{zaidan-etal-2007-using} uses an explain-predict method, where the explainer model first generates a human readable explanation. This explanation is then used as input to the predictor model on the actual task.

In detecting hallucinations, previous works \cite{zhou2021detecting} and section \ref{approach1} utilize a proxy model to help explain/detect hallucinations. Moreover, section \ref{approach1} allows for a more fine-grained interpretable detection. The novelty in the approach presented is that the extent to which the true reasoning of the model for faithfulness tag generation is reflected is increased.

The use of a single model is proposed that generates a summary token at timestep \emph{t} and predicts a faithfulness tag at that timestep \emph{t}. Since the generation of the token and prediction output tag are both simultaneously conditioned on the same computation graph, the model becomes more faithful in the task of detecting hallucinations.

\subsection{Design}
The design presented builds upon the \emph{seq2seq} transformer architecture detailed in \cite{vaswani2017attention}, maintaining its core structure of a bidirectional encoder and a left-to-right auto-regressive decoder. The proposed modifications aim to extend the architecture of Section \ref{apprch2-baselines} to better handle the dual tasks of summary generation and hallucination detection.

The structure of the model, abstracting the common parts in the baselines, is shown in Figure \ref{joint_model}. The input to the network is a dialog \emph{D}, which is a sequence of words \emph{$w_1$}, \emph{$w_2$}, ..., \emph{$w_T$}, where \emph{T} is the length of the dialog. The network consists of two outputs at every time-step \emph{t}: the summary token \emph{$o^s$} and the "faithfulness tag" \emph{$o^f$}. The modifications to the baseline model are described below.

The baseline model has a Language Model head, \emph{$L^s$} that projects the decoder hidden states, $\overrightarrow{\emph{$h_t$}}$ into a space, \emph{$y^s_t$} which generates the summary tokens after a softmax operation. Taking inspiration from \cite{zhang2016joint}, a linear classification layer, \emph{$L_f$} is proposed that projects the same decoder hidden states, $\overrightarrow{\emph{$h_t$}}$, into a "faithfulness" tag prediction space, \emph{$y^f_t$}. Since there are six tag categories, at every time-step \emph{t}, the \emph{$L^f$} will project \emph{$h_t$} into a dimension of six. The "faithfulness" tags \emph{$o^f$} are finally generated after a softmax operation.
Formally, 
\begin{equation}
    \label{summaryspace}
    y^s_t = W^s \overrightarrow{h_t} + b^s 
\end{equation}
\begin{equation}
    \label{summarytoken}
    o^s_t = softmax(y^s_t)
\end{equation}
\begin{equation}
    \label{tagspace}
    y^f_t = W^f \overrightarrow{h_t} + b^f 
\end{equation}
\begin{equation}
    \label{tag}
    o^f_t = softmax(y^f_t)
\end{equation}

where $W^s$ and $W^f$ are transformation matrices for token generation and tag prediction respectively, $b^s$ and $b^f$ are bias vectors.  

\subsection{Experiment Setup}

\subsubsection{Aligning a classification task layer to a generative model}

This approach involves attaching a classification head to a generative model. To achieve this, the source code of the BART python file from the HuggingFace Transformers library was altered by adding a linear layer that takes the decoder outputs as input and transforms them into a faithfulness classification space.

Additional problems were encountered in aligning the two heads - the language model head and classification head - together. These problems are listed below.

\begin{enumerate}
    \item Adding Target Labels for Added Faithfulness Classification Task.

    The original HuggingFace code for BART conditional generation forward function only allows for the input, attention mask, and labels. The labels here would be the target summary tokens. However, this approach makes it so there is now an extra target - faithfulness tags. To accommodate them, changes are made to the dataloader input as well as the forward function of the Trainer. To both, an extra field - \textbf{tags} - is added. Essentially, any fields that are in the train dataloader will be sent to the forward function if there is an identical definition of the same.

    During inference or generation, however, it's a different story. For inference, the input data needs to be sent through the trainer's prediction\_step function.
    
    \item Ensuring Compatibility of Special Tokens

    The embedding indices [1, 2, 3] for the BART tokenizer and model stand for special tokens. The index 1, especially is the $<PAD>$ token. Originally the faithfulness classes were originally in the range of 0-5, with class 1 standing for wrong reference or W. Because of this setup, the classifier would get confused on the actual semantic definition of the embedding index 1 - is it a PAD token or a class? Hence, the class range was shifted from 0-5 to 3-8. To accommodate this change, the output space of the classifier was also changed to 8.
    
    \item Ignoring the Pad Tokens for Inference
    
    The classifier would also predict/generate classes for the timesteps that the LM head would generate PAD tokens. To accomodate this, we introduce a mask, that only allows the classfier to predict on timesteps that are not special tokens on the LM head logits.
    
\end{enumerate}

\subsubsection{Returning both Summary Tokens and Faithfulness Tags}

In order to return both summary and faithfulness tags, a custom generate function had to be written which operates on both logits. This generate function involves a two-step approach. First, the summary logits are processed with logic identical to regular generate. Next, argmax is applied to the tag logits and then aggregated. 

\subsection{Training Strategy}

\begin{figure*}[h]
  \includegraphics[width=\textwidth]{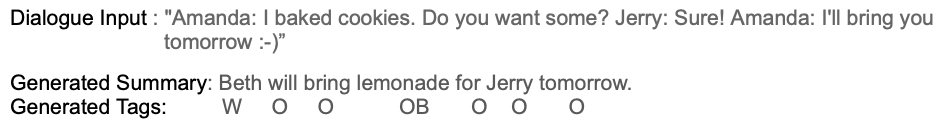}
  \caption{Ideal Model Output for Approach 2}
  \label{idealapprch2}
\end{figure*}

\subsubsection{Overview}

The training strategy is designed to ensure effective training of both the faithfulness classifier and the generative model. This involves a two-phase process focusing on different aspects of the model's capabilities.

\subsubsection{Phase 1: Training the Faithfulness Classifier}

\textbf{Objective:} To train the classifier to accurately identify and tag hallucinated content in the summaries. \\
\textbf{Dataset:} Utilizing the annotated MetaSAMSum dataset, which provides dialogues with corresponding summary tokens and faithfulness tags. \\
\textbf{Training Approach:} 
\begin{itemize}
    \item Isolated training of the classifier to focus on detecting hallucinations.
    \item Supervised learning with annotated faithfulness tags serving as ground truth.
\end{itemize}
\textbf{Expected Outcome:} The classifier should proficiently distinguish between faithful and hallucinated content in the summaries.

\subsubsection{Phase 2: Joint Model Training}

\textbf{Objective:} To train the entire joint model to generate coherent and accurate summaries while tagging hallucinated content. \\
\textbf{Dataset:} Training on the SAMSum dataset, offering a wide range of dialogues for summary generation. \\
\textbf{Training Approach:} 
\begin{itemize}
    \item Simultaneous training of the generative model and the classifier to enhance synergy.
    \item Use of a joint loss function to balance summary accuracy and hallucination detection.
    \item Fine-tuning the model for harmonious functioning of both components.
\end{itemize}
\textbf{Expected Outcome:} The model should generate high-quality summaries that are both coherent and faithful, with an enhanced ability to tag hallucinations accurately.

\begin{table}[h]
    \centering
    \begin{tabular}{|l|l|}
        \hline
        \textbf{Parameter}     & \textbf{Value} \\ \hline
        Learning Rate          & 2e-5           \\ \hline
        Train Batch Size       & 4              \\ \hline
        Eval Batch Size        & 1              \\ \hline
        Weight Decay           & 0.01           \\ \hline
        Number of Epochs       & 15             \\ \hline
        Seed                   & 42             \\ \hline
    \end{tabular}
    \caption{Training Hyperparameters}
    \label{table:training_hyperparameters}
\end{table}

\subsection{Evaluation}
\begin{figure*}[h]
  \centering
  \includegraphics[width=\textwidth]{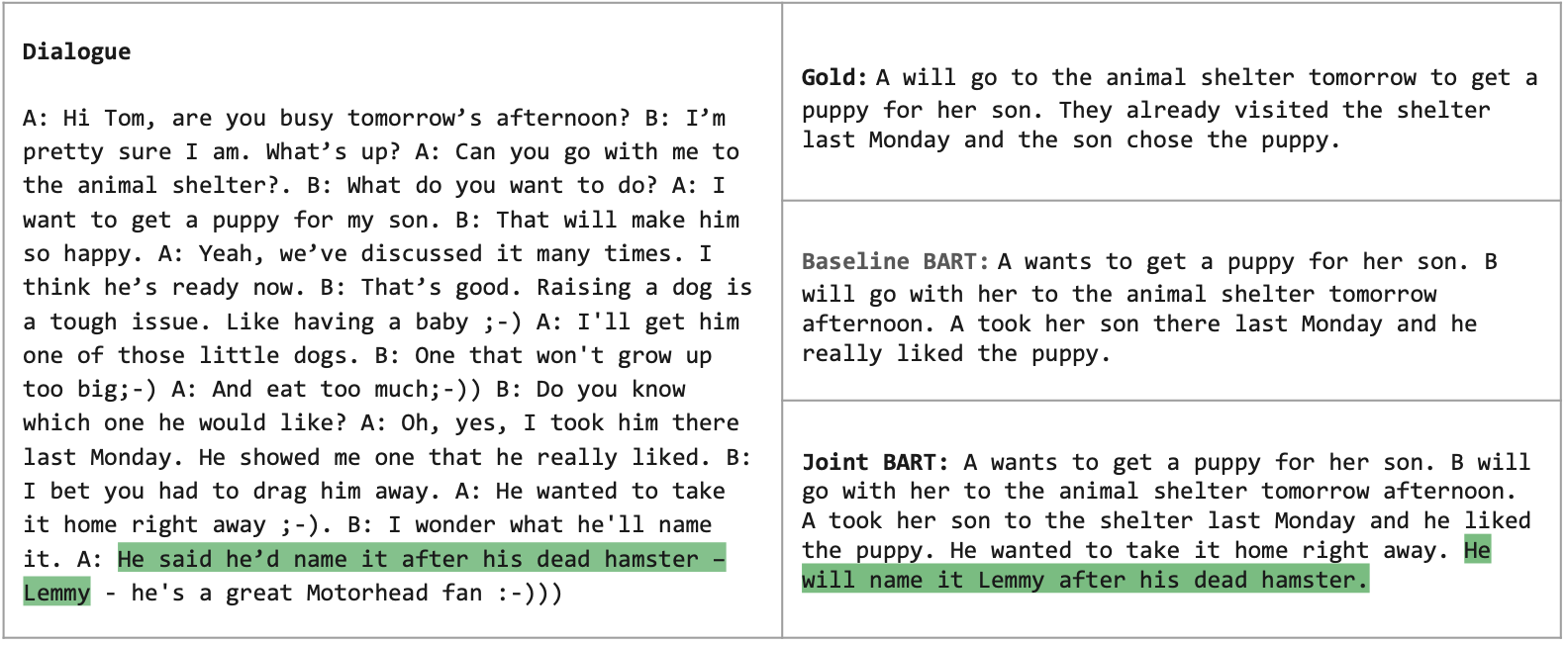}
  \caption{The summary generated by the joint model includes additional factual details that enhance the quality of the summary.}
  \label{interesting_result}
\end{figure*}

In this section, the ideal model output is defined as the primary metric of evaluation, and how the success of the hypothesis would be evaluated is described.

Figure \ref{idealapprch2} shows the ideal model, which consists of a summary and accurately identified hallucinated tokens with correct "faithfulness" tags assigned to them.

This approach consists of two parts: summaries and faithfulness tags. The evaluation consists of evaluating each part independently.

\begin{figure}[h]
  \centering
  \includegraphics[width=0.5\textwidth, height = 4.5cm]{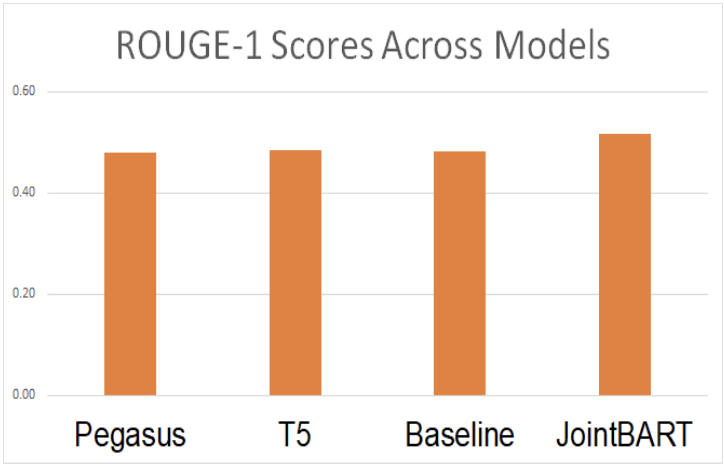}
  \caption{Joint Model Results}
  \label{joint_model_results}
\end{figure}

One of the main objectives is to maintain the summarization quality of the model. Hence, to capture this, existing metrics used to evaluate summaries - ROUGE metrics \cite{lin2004rouge} - are utilized.

For evaluating the model's performance in detecting and categorizing hallucinations by the use of the faithfulness tags, simple precision-based metrics used in token classification tasks would be used. However, since the joint model is fine-tuned, the summaries generated on the test set may not match the gold summaries and hence a gold set of "faithfulness" tags may not exist. For this reason, 10\% of the test set will be manually evaluated for the generated faithfulness tags.

\subsection{Results}

Upon evaluating the chart described in Figure \ref{joint_model_results}, the results of the proposed model compared to previous works and the baseline can be seen. Despite using only a small amount of data for fine-tuning the proposed model, a \textbf{+0.4 improvement in the ROUGE score} is observed compared to the other models.

Additionally, it was found that in some cases the proposed model generated better summaries than the gold summaries themselves. This is an interesting result considering the model was trained on limited examples.

\subsection{Discussion}

Upon human evaluation of the summaries generated by the joint model, an interesting and not uncommon pattern was noticed. In many cases, the summaries produced by the joint model contained more factual information than the human-written gold summaries and the baseline. An example is shown in Figure \ref{interesting_result}. As seen, the summary generated by the joint model includes additional factual details that enhance the quality of the summary. However, since ROUGE is a reference-based metric, the baseline-generated summary will incorrectly receive a higher score than the joint model's summary. This raises the question of whether a more robust metric is needed to measure summary quality.

\section{Conclusion}

This line of experimentation and analysis proposed by the paper marks a significant step towards hallucination detection and more faithful content generation. The major contributions are:

\begin{enumerate}

\item Dataset - In conclusion, a critical gap in existing datasets is addressed through the introduced nuanced approach to hallucination classification. Unlike traditional binary classification, token-level classification tags are employed in the presented dataset, providing deeper understanding of the nature and distribution of errors. "Missing Information" is highlighted as the most prevalent error, underscoring the significance of addressing missing information in future development.

\item LLM Experiments - The capability of LLMs such as GPT-4 and LLaMa2 regarding hallucination detection and summarization is demonstrated. Both models exhibit improved summary generation when explicitly tasked with detecting and explaining hallucinations.

\item Joint Model - A more accurate and faithful summarization is enabled through the presented joint model framework. Exploration of the MetaSAMSum data enabled a better-performing classifier, enhancing the model's ability to identify and handle hallucinations.
\end{enumerate}

In conclusion, current methodologies for hallucination detection are advanced through the presented research, providing a new direction for ongoing efforts toward faithful and factual content generation.

Future enhancements could include refining the joint loss mechanism and expanding the annotated MetaSAMSum for more effective training. Additional efforts could also focus on exploring evaluation methods beyond ROUGE Scores for a comprehensive assessment of hallucination detection.

\bibliography{main}
\bibliographystyle{acl_natbib}

\appendix

\end{document}